\documentclass[10pt, conference, compsocconf]{IEEEtran}
\usepackage{cite} 
\usepackage{amsmath,amssymb,amsfonts}
\usepackage{graphicx}
\usepackage{textcomp}
\usepackage{xcolor}
\usepackage{gensymb}
\usepackage{subfigure}
\usepackage{booktabs}
\usepackage{multirow}
\usepackage{marvosym}
\usepackage{stfloats}
\usepackage{graphicx}                                                        
\usepackage{float} 
\usepackage{comment}
\usepackage{verbatim}
\usepackage{algorithm}
\usepackage{hyperref}
\usepackage{algorithmicx}
\usepackage{algpseudocode}
\usepackage[numbers,sort&compress]{natbib}
\ifCLASSINFOpdf
\else
\fi
\begin{document}

\title{Path Planning for Air-Ground Robot Considering Modal Switching Point Optimization}
\author{
\IEEEauthorblockN{1\textsuperscript{st} Xiaoyu Wang}
\IEEEauthorblockA{The School of Vehicle and Mobility\\
Tsinghua University\\
Beijing, P.R.China\\
Email: 273456@whut.edu.cn}

\and
\IEEEauthorblockN{2\textsuperscript{nd} Kangyao Huang}
\IEEEauthorblockA{The Department of \\
Computer Science and Technology\\
Tsinghua University\\
Beijing, P.R.China\\
Email: huangky22@mails.tsinghua.edu.cn}

\and
\IEEEauthorblockN{3\textsuperscript{rd} Xinyu Zhang*{\Letter}}
\IEEEauthorblockA{The School of Vehicle and Mobility\\
Tsinghua University\\
Beijing, P.R.China\\
Email: xyzhang@tsinghua.edu.cn}

\and
\IEEEauthorblockN{4\textsuperscript{th} Honglin Sun}
\IEEEauthorblockA{The School of Vehicle and Mobility\\
Tsinghua University\\
Beijing, P.R.China\\
Email: hsun@akane.waseda.jp }
\\
\IEEEauthorblockN{7\textsuperscript{th} Jun Li}
\IEEEauthorblockA{The School of Vehicle and Mobility\\
Tsinghua University\\
Beijing, P.R.China\\
Email: lj19580324@126.com }

\and
\IEEEauthorblockN{5\textsuperscript{th} Wenzhuo Liu }
\IEEEauthorblockA{The School of Vehicle and Mobility\\
Tsinghua University\\
Beijing, P.R.China\\
Email: liuwenzhuo2023@163.com}
\\
\IEEEauthorblockN{8\textsuperscript{th} Pingping Lu}
\IEEEauthorblockA{University of Michigan,Ann Arbor, USA\\
 Email: pingping\underline{~}lu0913@126.com}

\and
\IEEEauthorblockN{6\textsuperscript{th} Huaping Liu}
\IEEEauthorblockA{The Department of\\
Computer Science and Technology\\
Tsinghua University\\
Beijing, P.R.China\\
Email: hpliu@tsinghua.edu.cn }

}
\maketitle

\begin{abstract}

An innovative sort of mobility platform that can both drive and fly is the air-ground robot. The need for an agile flight cannot be satisfied by traditional path planning techniques for air-ground robots. Prior studies had mostly focused on improving the energy efficiency of paths, seldom taking the seeking speed and optimizing take-off and landing places into account. 
A robot for the field application environment was proposed, and a lightweight global spatial planning technique for the robot based on the graph-search algorithm taking mode switching point optimization into account, with an emphasis on energy efficiency, searching speed, and the viability of real deployment. 
The fundamental concept is to lower the computational burden by employing an interchangeable search approach that combines planar and spatial search. 
Furthermore, to safeguard the health of the power battery and the integrity of the mission execution, a trap escape approach was also provided. Simulations are run to test the effectiveness of the suggested model based on the field DEM map. The simulation results show that our technology is capable of producing finished, plausible 3D paths with a high degree of believability. Additionally, the mode-switching point optimization method efficiently identifies additional acceptable places for mode switching, and the improved paths use less time and energy.

\end{abstract}

\IEEEpeerreviewmaketitle

\section{Introduction}

In recent years, new kinds of flying and driving robots have evolved as the theory and technology of electric vertical take-off and landing have advanced. For more flexible transportation, scientists and engineers combine electrically propelled chassis technologies with aviation applications to create hybrid locomotion air-ground vehicles. Great achievements have been made and attracted widespread attention, such as TF-X from Terrafugia\cite{dietrich2011roadable}, AAV from EHang\cite{du2017multi}, the EPFL jumpglider from Carnegie Mellon University and Swiss Federal Institute of Technology in Lausanne\cite{M2009Towards,2011The}.
The combined flying and driving skills, as well as the ability to shift power from propellers to wheels for environmental adaptation, are common elements of these breakthroughs. The focus of robotics is on this capacity for multimodal movement. 
 However, when energy consumption is taken into consideration, cruising in the sky requires high-power density to overcome gravity which significantly limits the endurance and task scope\cite{2008Flight}.
To balance driving and flying, the right path planner for 3D searching and navigating algorithms become essential.
\par There are mainly sampling-based algorithms, graph-search algorithms, curve interpolation methods, etc. \cite{zammit2018comparison,2012Advanced,2016Optimal} for 3D path planning. Due to the increasing search volume in 3D scenarios and the requirement for fast planning of UAV\cite{2014Fast}, sampling-based algorithms, especially RRT \cite{1998Rapidly} and its variants (such as RRT-Connect \cite{2000RRT} and RRT* \cite{2011Sampling}), are still the mainstream. However, when it comes to air-ground vehicles, modal switching should be considered.
\par In the category of graph-search algorithms, A* and its derivation algorithms are also widely used in 3D scenarios\cite{2020Path}. Incremental searching algorithms such as LPA* \cite{2004Lifelong} and D* Lite \cite{Koenig02d*lite} are able to dynamically search a path in a partially known environment. ARA* \cite{Likhachev05anytimedynamic} can rapidly generate a sub-optimal path and iteratively optimize by adjusting the weight of the heuristic function. Theta* \cite{Nash07theta*:any-angle} implements an any-angle path planning through the process of removing parent nodes. HPATheta* \cite{CHAGAS2022116061} 
implements the hierarchical path planning method, which can complete the large-scale virtual terrain pathfinding, and smooth the path within a reasonable calculation time.
\par The path planning methods for ground robots and aerial robots are different. The hybrid robots are able to fly so that planar planning methods are no longer suitable\cite{2011Probabilistic}. While employing the planning method for aerial robots will be unable to make use of the high energy efficiency feature from ground driving. Despite graph-search based 3D path planning algorithms walks through fair development, the planning method for the air-ground robot still stays in a primary stage\cite{2016A}. Nevertheless the multi-modal movement feature of the air-ground robot brings new challenges\cite{Roberge2013Comparison}\cite{9839587}.
\par In this field, Brandon Araki et al. presented SIPP and ILP algorithms for hybrid swarm robots \cite{7989657} which can be used on multiple robots in the continuous-time domain. Meanwhile, this algorithm also focuses on time and energy consumption for future 3D traffic networks. Amir Sharif et al. presented an energy-efficient A* based method implemented by modifying the calculation of heuristic cost for different actions \cite{2018Energy}. H.J. Terry Suh et al. presented the first motion planning method for the multi-modal locomotion system \cite{2019arXiv190910209S}, which considered energy efficiency and dynamic constraints. All these works presented viable planning methods for the air-ground robot with high energy efficiency but barely mentioned the computation time, real-time obstacle avoidance when the transition between flight and ground driving is more barely mentioned.

\section{Background and related work}
\par A path planning method for air-ground robot  based on a hybrid A* and BAS (Beetle Antennae Search) algorithm to achieve a balance between flying and driving was proposed, it can switch modes shrewdly in challenging field conditions, and produce task execution routes with minimal overall energy consumption. There are mainly four contributions of our works:

\begin{itemize}
    \item Introducing the novel variable-structure air-ground robot which is capable of operating in challenging field environment. The folding arm of the robot can make the robot structure compact and flexible, and enhance the maneuvering performance in the field environment.
    \item An iterative framework for path optimization is proposed to achieve the balance between computing time and accuracy, 2D searching and 3D searching operate alternatively until the global optimum is reached.
    \item Improved BAS optimization algorithm is proposed inserting the modification algorithm into the real-time computational logic of path planning, to optimize the takeoff and landing node. Not only to save energy consumption, but also increase the feasibility of the calculated route, and enhance the safety of the path for the robot field rescue mission.
    \item The proposed path planning model can deal with the development of various huge slopes and emergency threats through the node information judgment logic in the field environment. It can also take into account the health and mobility of the robot's power system.
\end{itemize}

\par The remainder of the paper will proceed as follows:  Section \uppercase\expandafter{\romannumeral2} introduces the robot platform our team designed for this experiment. In Section \uppercase\expandafter{\romannumeral3}, the structure and detail of the proposed path planning method is described. Section \uppercase\expandafter{\romannumeral4} presents the simulation  experiment designed to verify the effectiveness of the proposed algorithm framework, and includes simulation results. Finally, Section \uppercase\expandafter{\romannumeral5} reviews the conclusions made from  simulation results and future work.

\section{Robot Platform}

\par
\begin{figure}[t]
    \centering
    \includegraphics[width=0.44\textwidth]{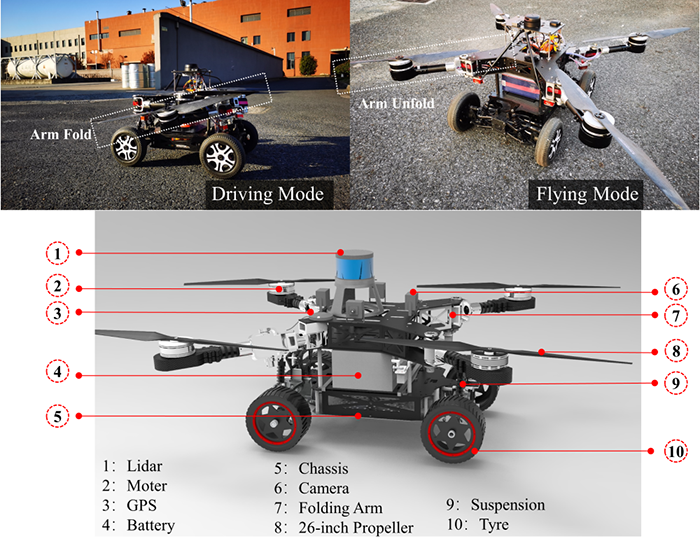}
    \setlength{\abovecaptionskip}{0.cm} 
    \setlength{\abovecaptionskip}{0.cm}
    \caption{The overall structure of the proposed robot}
    \vspace{-6mm}
    \label{fig1}
\end{figure}

 In our previous works, several types of air-ground robots are designed to carry workload in rescue and searching\cite{Tan2021a}\cite{zhang2022multi}. These robots have achieved good results in testing under urban scenarios, even in some complex fields such as urban rescue and logistics. To get a better performance of driving and passability, the innovative robot in \cite{Tan2021a} is upgraded by folding propellers. Our platform are designed with two modals: the open-wing state for flying in the air, and the wing-folded state for driving on the ground. The two modals and main hardware compositions of the robot are shown in Fig. \ref{fig1}. Our robot is expected to be fully self-organized and intelligent to perform high-risk works replacing human. 

\begin{figure*}[t]
    \vspace{-0.8cm}
    \centering
    \includegraphics[width=1\textwidth]{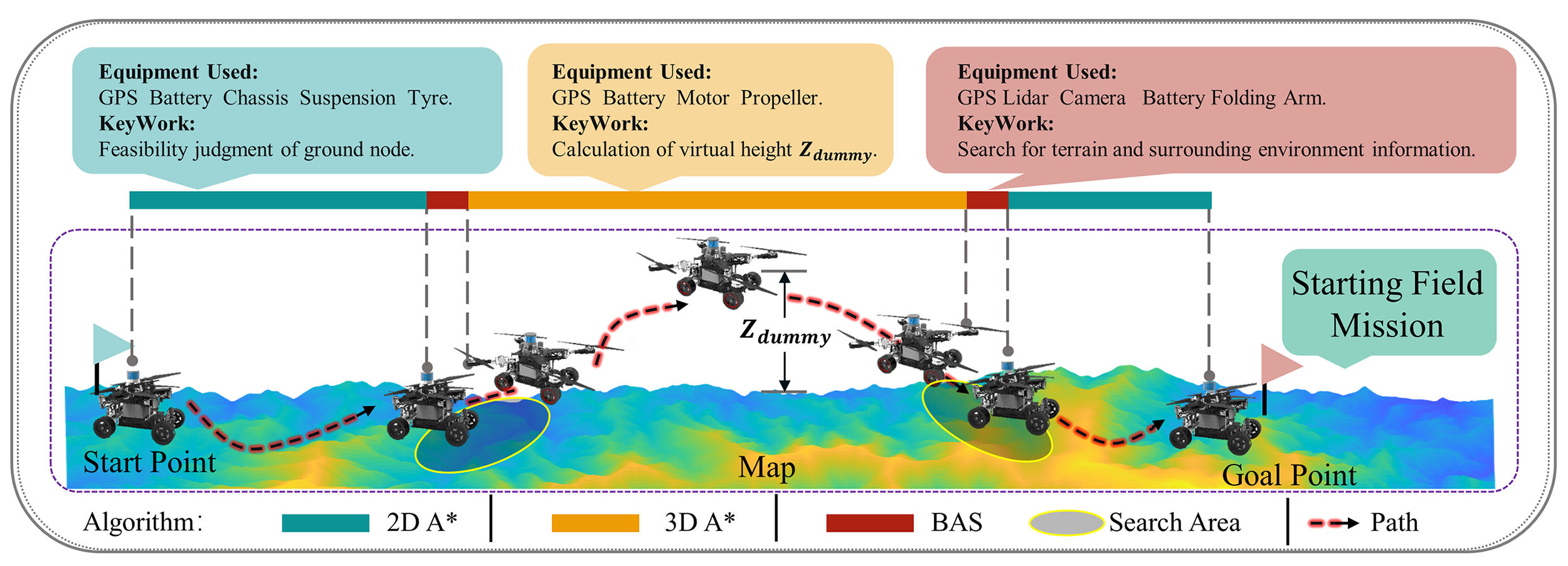}
    \vspace{-4mm}
    \caption{The operation diagram of the proposed algorithm. (The equipment used are the primary hardware that the robot  employs in each of the several operating modes shown in Fig. 1. The keywork is the main task in the process of corresponding algorithm iteration)}
    \setlength{\abovecaptionskip}{-0.2cm}
    \vspace{-6mm}
    \label{figzhengti}
\end{figure*}

\section{Method}

\par
\begin{figure}
    \centering
    \includegraphics[width=0.4\textwidth]{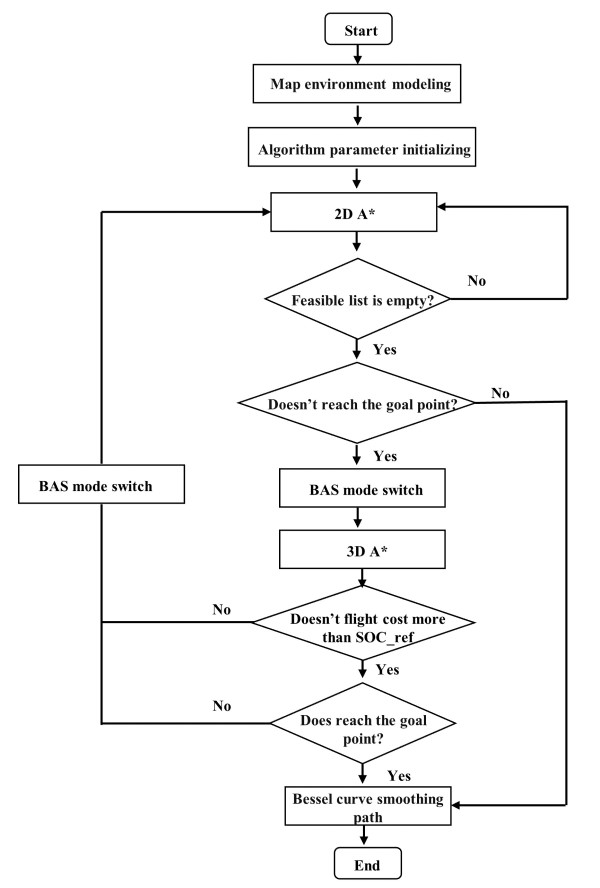}
    \caption{Flowchart of the planner execution}
    \vspace{-6mm}
    \label{figliucheng}
\end{figure}

 The proposed planning algorithm is described in detail in this section. This approach of planning is typically based on the characteristics of air-ground robots: when driving on the ground, a robot has good energy efficiency, limited mobility, low energy consumption, and low computational load; when flying in the air, the robot experiences the opposite outcome. Therefore, we assume that the robot typically travels on land and only flips to flying mode when it becomes difficult to identify a path or when the terrain is insufficient for the air-ground robot's mobility. The challenge is to enable the robot to complete the rescue operation while using less time and energy, and to do it while simultaneously deciding when to take off and land.
By choosing a single mode switching point that is inconvenient for the platform's takeoff and landing, the A* algorithm can result in hazardous scenarios like roll and sideslip. These problems can be avoided by using the BAS algorithm.
The planner will alternately execute the 2D and 3D path finding algorithms of the BAS mode switching point optimization method given a start and target position. Following several iterations, the planner will produce a raw path with a greater energy efficiency. The route is then refined using Bessel curve smoothing to gauge the robot's mobility. The robot begins its field rescue mission after reaching the target point. In summary, the overall flow of the proposed algorithm to complete the task is shown in Fig. \ref{figzhengti}. The flowchart in Fig. \ref{figliucheng} shows the basic logic of the proposed path planner. 

\subsection{Path Planning}

\subsubsection{Ground Planning: 2D A*}

\par The planner searches in a plane according to the original A* cost function when the robot stays on the ground. Meanwhile, the function to judge the feasibility function of air-ground robot driving in terrain (called takeoff decision function) runs together with A*. Takeoff decision function can determine whether the maneuverability index(M-index) of the robot is greater than the slope, flip angle and turning radius(DOF) from the current node to the next node. Based on the maneuverability of air-ground Robot, the experimental tests to obtain empirical values for the maneuverability index was designed. The function will count a number (count) from 0 if the DOF between the current node and adjacent nodes is greater than or equal to M-index until the count exceeds the threshold (thre).And move the node from the open list to the close list which do not match M-index. Otherwise, a new node will be recorded and the count will be reset to 0.When count = 7, the takeoff decision function returns the flag with true value, the planner stops searching and picks current node  as the initial mode switching point.
\par In addition to the Manhattan distance in the conventional A* algorithm, the suggested method's cost function additionally accounts for the energy consumption $E$ in \nameref{Energy Consumption Model} between nodes in the field environment.
 \par According to the mobility test experiment of the air-ground Robot, between the feasible node and the previous node shall meet the following conditions:

\floatname{algorithm}{Algorithm}

\begin{algorithm}
    \caption{Take off Decision Function}
    \begin{algorithmic}[1]
    \Function{}{$DOF, M-index, count, flag$}
        \If{$DOF>M-index$}
            \State $count++$
        \Else
            \State $count \gets 0$
        \EndIf
        \If{$count>(thre=7)$}
            \State $flag \gets true$
        \Else
            \State $flag \gets false$
        \EndIf
        \State \Return {$DOF, M-index, count, flag$}
    \EndFunction
    \end{algorithmic}
\end{algorithm}
\begin{equation}
\begin{cases}
gx_{\text{min}} \leq gx \leq gx_{\text{max}}\\
gy_{\text{min}} \leq gy \leq gy_{\text{max}}\\
gz_{\text{min}} \leq gz \leq gz_{\text{max}}
\end{cases}
\end{equation}
\subsubsection{Flight Planning: 3D A*}

\par According to the distance to the goal, the 3D pathfinding method was categorized into two categories : flying to the goal and flying to the ground. We expect that when the target is close, the robot would fly straight there, ignoring the goal's altitude, as often switching modes might lead to instability and energy waste. The 3D A* was utilized to look for a path that leads to the desired outcome. To preserve energy, the robot was expected to fly as long as it can if the destination is far away. In this scenario, a modified 3D A* was employed to enable the robot to fly away from the present seeking trap, called a trap escape algorithm in short.
\par The trap escaping algorithm is based on the assumption that the robot can across over current terrain obstacles. In this algorithm, a variable $H_{2D}$ and an SOC (State of Charge) judement rule was introduced.
\par \textbf{$H_{2D}$:} The horizontal Manhattan distance from the current position to the goal, same as the heuristic cost in the 2D pathfinding phase. $H_{2D}$ is used as an indicator of whether the robot is approaching the goal. 
\par At the beginning, an initial value named $H_{2D0}$ was recorded, and which should also be equal to the final value of $H_{min}$ in the 2D pathfinding phase. In each execution loop, calculate $\Delta H_{2D}$, which represents the moved distance toward the goal. 
Next, depending on the value of $\Delta H_{2D}$, 3 stages of the 3D pathfinding process were defined : 1. takeoff stage; 2. escape stage; 3. landing stage.
\par In the trap escaping algorithm, a dummy altitude variable $z_{dummy}$  we introduce, and the values according to the current stage to let the searching runs as desired were assigned.

\par Concretely, the takeoff stage is to demand the robot fly upward until it can horizontally get closer to the goal. When $\Delta H_{2D}<C_{escape}$, the dummy altitude $z_{dummy}$ equals to the current robot height $z_{curr}$ added with a small positive value $\epsilon$. The dummy altitude is applied to calculate the heuristic value instead of the actual altitude of the goal. Since the original heuristic function would lead to the searching direction attracted to the same altitude with the goal, while $z_{curr}$ is used to cancel this effect, and the positive $\epsilon$ drives the algorithm searching upwards. Otherwise, when the goal is also on the ground, the attraction effect of the heuristic function would result in meaningless searches near the ground. When $C_{escape}\leq \Delta H_{2D}<C_{landing}$, the planner will get into the escape stage when the robot is flying over the obstacle. The value of dummy altitude are required to be set the same value as the current robot height to eliminate the effect of height, which would lead the algorithm to search forward. Finally, as the $H_{2D}$ gets lower, when $\Delta H_{2D}\geq C_{landing}$, the dummy altitude equals were set to the ground height for attracting the searching direction to the ground when considering the robot has gotten rid of the obstacle. Once a ground node is searched, planner will record it as the landing point, and reconstruct the path then terminate the loop. 

\par \textbf{SOC judement rule:} 
When the platform departs from the closest mode changeover point, the SOC record begins. The mode switch condition will be entered if the overall soc consumption of the battery during the flight exceeds the pre-set safety limit ($SOC_{ref}$). The regulation is intended to keep the battery soc in an effective and healthy range while in flight and to guarantee that field rescue reconnaissance missions may be carried out without endangering the battery's long-term health.

\floatname{algorithm}{Algorithm}
\begin{algorithm}[t]
    \caption{$H$ Calculation for the Trap Escaping Algorithm}
    \begin{algorithmic}[1]
        \If{$H_{2D0}<H_{2D}$}
            \State $H_{2D0} \gets H_{2D}$ \text{\hspace{16mm}// update $H_{2D0}$}
        \EndIf
        \State $\Delta H_{2D} \gets H_{2D0}-H_{2D}$
        \If{$\Delta H_{2D}<C_{escape}$}
            \State $z_{dummy} \gets z_{curr}+\epsilon$ \text{\hspace{7mm}// takeoff stage}
        \ElsIf{$C_{escape} \leq \Delta H_{2D}<C_{landing}$}
            \State $z_{dummy} \gets z_{curr}$ \text{\hspace{12mm}// escape stage}
        \ElsIf{$\Delta H_{2D}>C_{landing}$}
            \State $z_{dummy} \gets z_{ground}$ \text{\hspace{9mm}// landing stage}
        \EndIf
        \State $H \gets H_{2D}+|z_{dummy}-z_{curr}|$
        \If{$\Delta SOC_{flying}> SOC_{ref}$}
         \State $z_{dummy} \gets z_{ground}$ \text{\hspace{9mm}// landing stage}
        \EndIf
    \end{algorithmic}
\end{algorithm}

\begin{figure}[t]
    \centering
    \includegraphics[width=0.4\textwidth]{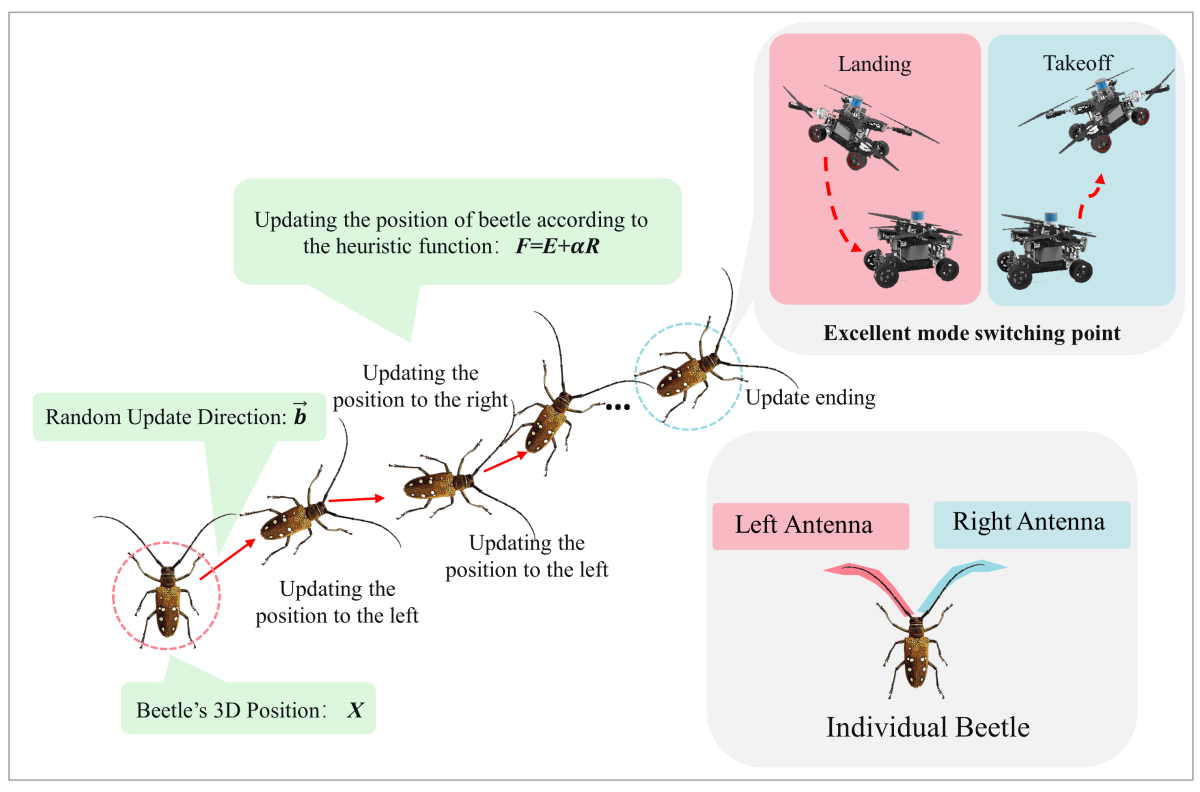}
     \vspace{-4mm}
    \caption{Schematic diagram of the BAS modal switching point optimization}
    \vspace{-4mm}
    \label{Bas}
\end{figure}

\subsection{Modal Switching Point Optimization} 

The principle of BAS algorithm is shown in Fig. \ref{Bas}.
\par According to the field application scenario of the robot, the main steps of BAS modal switching point optimization algorithm are as follows:
\par Step 1: The updating step direction of individual longicorn is random, and the random vector is expressed as follows:
\begin{equation}
\overrightarrow{b}=\frac{Rands(k,1)}{\left\|Rands(k,1)\right\|}
\end{equation}
\emph{Where, k is the spatial dimension, which is taken as 3 in this paper. The three dimensions respectively represent the longitude, latitude and height of nodes.}
\par Step 2: Set the initial mode switching point determined by 2D and 3D A* as the initial position of Longhorn.
\par Step 3: Longhorn updates the position of the left and right two tentacles:

\begin{equation}
\begin{cases}
X_{\text{R}}=X+D \frac{\overrightarrow{b}}{2}\\
X_{\text{L}}=X-D \frac{\overrightarrow{b}}{2}
\end{cases}
\end{equation}
\emph{Where, $X_R$ and $X_L$ are the coordinates of the left and right tentacles of the longhorn respectively. b is the coordinates of the centroid of the longhorn. D is the distance between two tentacles.}
\par Step 4: Based on the current position of the left and right tentacles, the size of fitness function F ($X_R$) and F ($X_L$) is compared to update the position of the centroid:
\begin{equation}
X=X- \delta \overrightarrow{b} sign (F(X_{\text{R}})-F ((X_{\text{L}}))
\end{equation}
\begin{equation}
F=E+\alpha R 
\end{equation}
\begin{equation}
R=a \cdot gx+b \cdot gy +c \cdot gz 
\end{equation}
\emph{Where, $\delta$ is the Euclidean distance of the step, E is energy consumption, $\alpha$, a, b and c are hyperparameter, gx, gy and gz are the gradient values of x, y, and z with the current node.}
\par The $\alpha$ values in different situations to select different weight relations between E and R  can be adjust. For example, to enable the BAS algorithm to search for mode switching points in a longer range, when the SOC of robot's power battery is sufficient the $\alpha$ value can appropriately increased.
\par Step 5:
The modal switching point data of each iteration is retained, and the node with the smallest fitness function is selected as the starting point of the next stage of planning.

\subsection{ Energy Consumption Model \label{Energy Consumption Model}}

\begin{table}[t]
 \setlength{\belowcaptionskip}{-4pt}
 \small
 \caption{Parameters Settings(supplement)}
 \setlength{\belowcaptionskip}{-4pt}
    \label{tab:Parameters Settings}
    \centering
    \begin{tabular}{@{}lll@{}}
    \toprule
    Symbol     & Meaning                    & Value                       \\ \midrule
    $\rho$     & Air density                & 1.2 kg/m$^3$ \\
    $m$        & robot mass               & 39.5 kg                     \\
    $r$        & Propeller radius           & 0.4191 m                    \\
    $x$        & Number of propellers       & 6                           \\
    $g$        & Gravitational acceleration & 9.81 m/s$^2$ \\
    $\eta$     & Motor efficiency      & 0.58                        \\
    $\mu$      & Coefficient of friction    & 0.06                        \\
    $C_d$      & Coefficient of air drag    & 1.5                         \\
    $v(\text{fly})$   & Flying velocity            & 2 m/s                       \\
    $v(\text{drive})$ & Driving velocity           & 1 m/s                       \\
    $A(\text{fly})$   & Flying windward area       & 0.6 m$^2$    \\
    $A(\text{drive})$ & Drving windward area       & 0.05 m$^2$   \\
    Constant   & Standby energy             & 100 J                       \\ \bottomrule
    \end{tabular}
   
    \end{table}

\par To evaluate the energy efficiency of a path, equations below are used to estimate energy consumption.
\begin{equation}
E=E_{\text{hover}}(mode)+E_{\text{move}}(mode)+E_{\text{transform}}(mode)
\end{equation}
\par The total energy consumption was assumed consists of hovering energy\cite{2018Energy}, moving energy, and transform energy.  $mode$ represents the operating mode of the robot.

\begin{equation}
E_{\text{hover}}(mode)=
\begin{cases}
\sqrt{\frac{1}{2\pi \rho}} \left(\frac{mg}{x}\right)^\frac{3}{2} \frac{x}{r} \frac{\Delta d}{\eta v} &mode=\text{fly}\\
\text{Constant} &mode=\text{drive}
\end{cases}
\end{equation}

\begin{equation}
E_{\text{move}}(mode)=
\begin{cases}
mg\Delta h + \frac{\rho AC_dv^2 \Delta d}{2} &mode=\text{fly}\\
\mu mg\Delta d + \frac{\rho AC_dv^2 \Delta d}{2} &mode=\text{drive}
\end{cases}
\end{equation}
\begin{equation}
E_{\text{transform}}(mode)=E_{\text{expand/fold}}+E_{\text{Bodeneffekt}}
\end{equation}

\par When hovering, the weight of the air blown downward by propellers is equivalent to its self-weight\cite{leishman2002principles}. In our estimation model, the moving energy consumption includes two parts: to overcome gravity or ground friction; and to overcome air drag. Hovering consumes the most energy. Hence the calculation of air drag was simplified by using fixed $C_d$ and $A$ for any direction and apply a fixed $\mu$ for ground moving.
The $E_{\text{transform}}(mode)$ denotes the energy consumption during the transform process. In this paper, when switching modes, the energy used to open or fold the wings is the $E_{\text{expand/fold}}$. And the energy consumed by the ground effect is the $E_{\text{Bodeneffekt}}$. The values are obtained by testing the actual working process of the robot. 
\par SOC calculation formula of power battery of air-ground robot:
\begin{equation}
SOC_{\text{t}}=\frac{Q_{\text{0}}-\sum_{0}^{t}E_{\text{t}}}{Q}
\end{equation}
\par Values of parameters used in the equations are shown in the table. \ref{tab:Parameters Settings}.

\begin{figure}[t]
    \centering
    \includegraphics[width=0.5\textwidth]{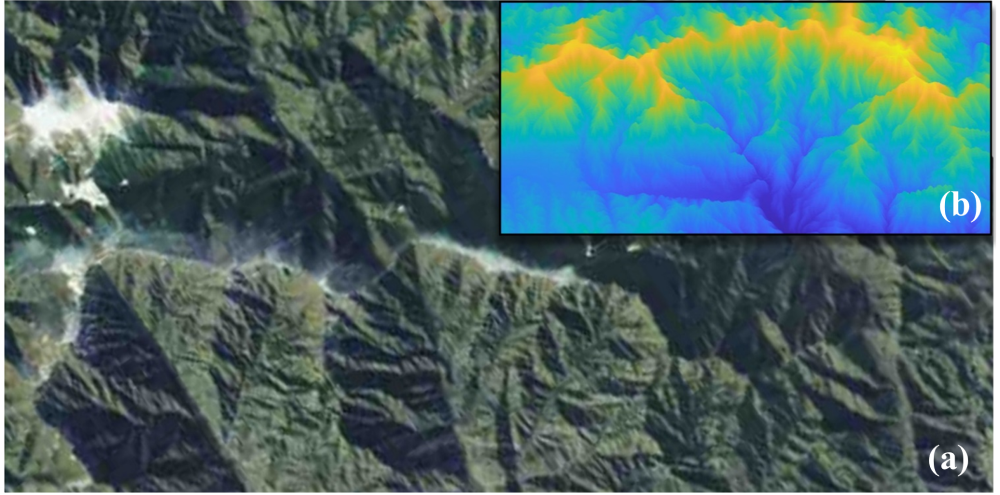}
    \vspace{-4mm}
    \caption{Map: (a) is the original satellite map of the experimental area and (b) is the map of the digital elevation data after $matlab$ processing }
    \vspace{-4mm}
    \label{weixingmap}
\end{figure}

\section{Simulation Results}

\begin{figure}[t]
    \centering
    \includegraphics[width=0.5\textwidth]{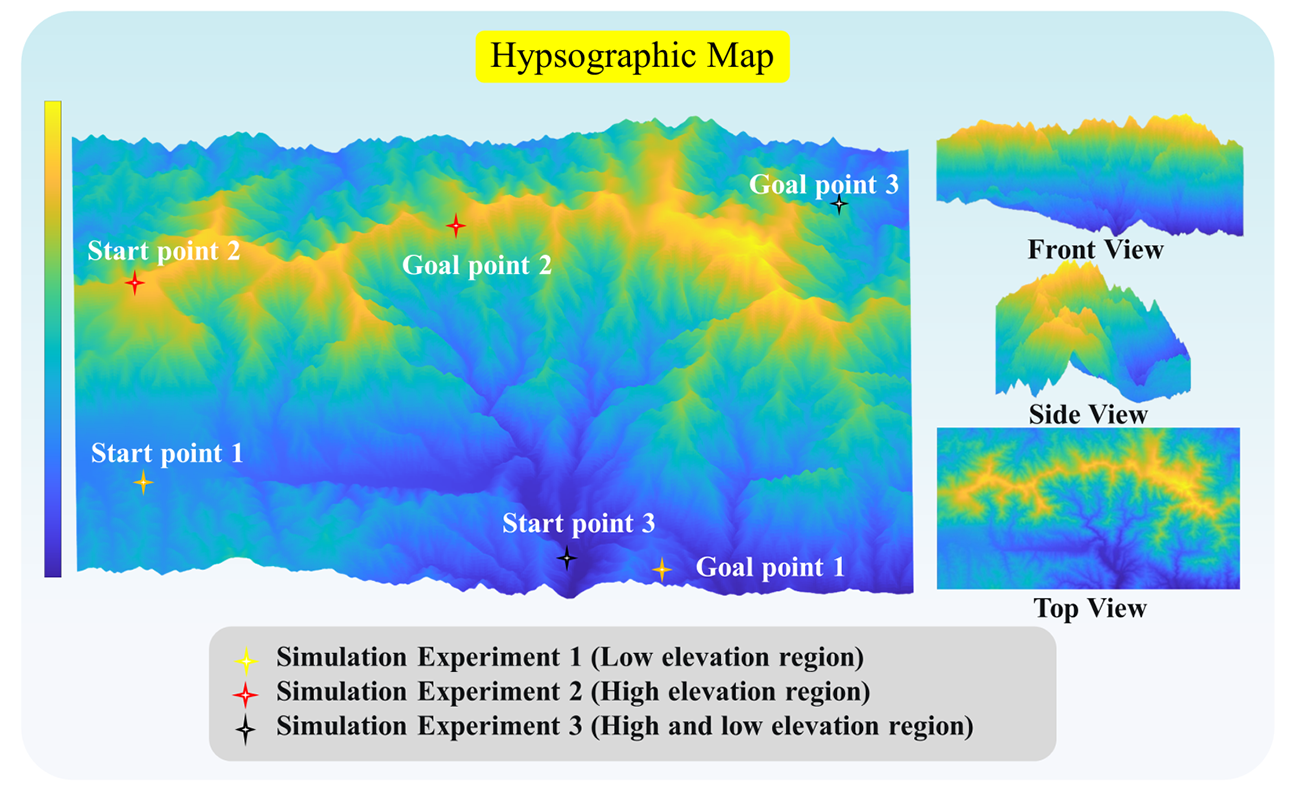}
    \caption{Distribution of start-goal points of the experimental setup and different angle views of the map }
    \vspace{-7mm}
    \label{figcontourmap}
\end{figure}

In order to verify the feasibility of the algorithm, the DEM (Digital Elevation Model) data map with 12m resolution is selected as the simulation scene. Mountainous and hilly landscapes with a wide range of altitudes are present in the selected map region, as seen in Fig.\ref{weixingmap}.
This simulation is based on the scene setting of a land and air robot mission execution, and it is separated into low altitude area, high altitude area, and composite region for experimental verification. The set starting and goal points in the three areas are shown in Fig. \ref{figcontourmap}. 

\subsection{Analysis} 

\begin{figure*}[t]
    \centering
    \includegraphics[width=1\textwidth]{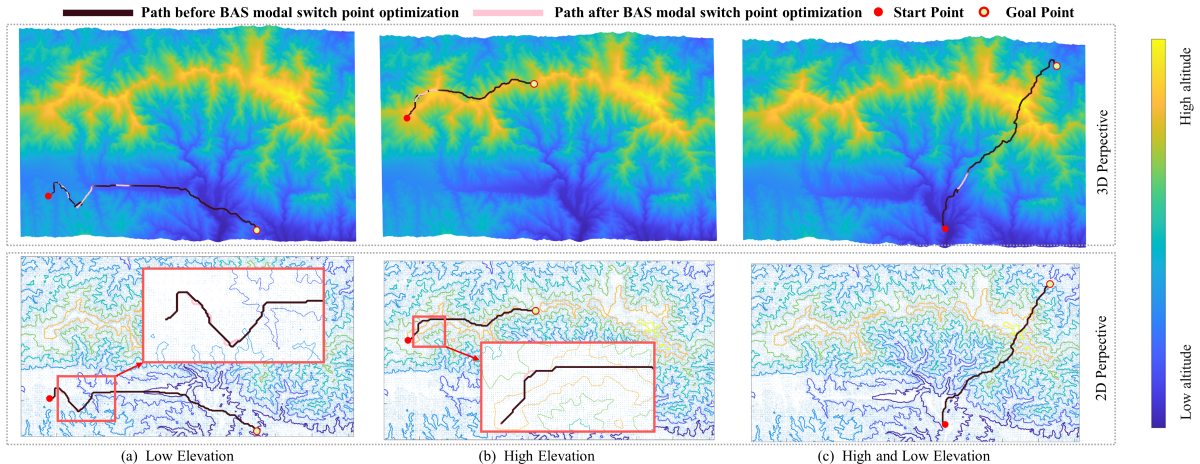}
    \caption{Path planning results of three elevation experiment}
    \label{figpath}
\end{figure*}

Fig. \ref{figpath} displays the experimental results of path planning in three distinct altitude situations. As can be observed, the algorithm is capable of adapting to a variety of field conditions and producing robot-friendly routes. The path planning impact without modal switching points is also extremely reasonable, and path finding can be carried out along the edges of contour lines. Since the height span is not very wide, we may discover the area with gradual altitude shift as the alternative area for the path, as shown in the low altitude and high altitude trials, (a) and (b) of Fig. \ref{figpath}, when paired with the 2D and 3D perspective data pictures.
While the high altitude lines generally go through the bright yellow high altitude region, the low altitude paths mostly travel through the dark blue low altitude region. Since the two regions' colors are essentially the same, the landscape created by the produced path matches the robot's movements more closely.

\par Since the terrain is more challenging to drive on the ground in high altitude regions, the planner uses flight mode more frequently to advance, and the soc consumption in this case is also the highest, as can be seen from the figure. The difference in elevation between the routes in high and low altitude regions is also greater. The path and direction in the two-dimensional image are unaffected by the optimization of the modal switching point of the BAS algorithm; only the robot's takeoff time is altered.
It is also quite flexible when switching between flight and ground forms; the suggested algorithm's recommended flow is followed by the switching operations of the 2d, 3d, and BAS algorithms.
The initial modal coordinate points are chosen when the BAS algorithm is applied based on the A* algorithm, and the information of the local real-time environment may be gathered and modeled using Lidar, cameras, etc. Different modifications are made to the routes both before and after the modal conversion. For robot takeoff and landing, the optimal sites have gradient values that are more advantageous.
The optimized task paths not only have shorter task execution time, but also have less power consumption, with 2.1$\%$, 2.5$\%$ and 2.2$\%$ of power consumption saved respectively in the three experiments.

\par 
Three traditional path planning search algorithms are provided as a comparison in order to evaluate the performance of the modal switching point method suggested in this article. The comparative performance of the four algorithms with the same number of search points is displayed in Fig. \ref{Performance}. Due to its simple iterative logic and cost function, the BAS algorithm has the quickest search time among the four algorithms. The parameter $R$ is employed in the aforementioned research to fully quantify the gradient strength of the modal switching sites, and the modal switching points chosen by the BAS algorithm had lower average $R$ values. Additionally, the entire path with the improved modal switching point uses least energy.

   
\section{Conclusion and Future Work}
In this study, an air-ground robot was developed that can withstand the challenging field conditions. Additionally, we primarily proposed a quick and lightweight spatial path planning method for the robot that incorporated a 2D/3D searching approach and a BAS modal switching point optimization algorithm. The method was tested successfully in simulated flight. To confirm the efficiency and viability of the proposed BAS algorithm, the proposed path planning framework can efficiently minimize the overall energy consumption compared to three different search algorithms.Therefore, Our technology may be used on air-ground robots, particularly those small-sized, highly mobile robots with rescue and reconnaissance capabilities. 
\par For future work, we will  carry out with two  aspects: we generally intend to conduct more real-world trials and advance the air-ground robot in the application. On the other hand, we will utilize our algorithm in completely unpredictable and dynamic contexts because that is more applicable to real-world situations.

\begin{figure*}
    \centering
    \includegraphics[width=0.8\textwidth]{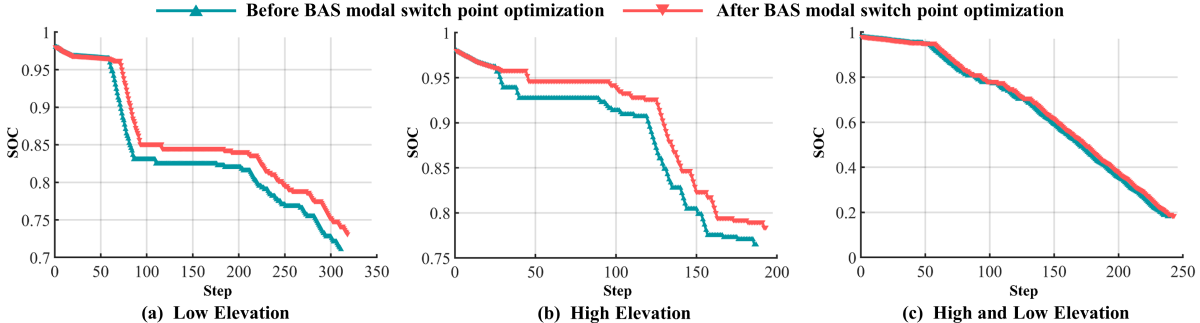}
    \caption{Comparison of SOC trajectories}
    \vspace{-4mm}
    \label{fig2}
\end{figure*}
  \begin{figure}
    \centering
    \includegraphics[width=0.5\textwidth]{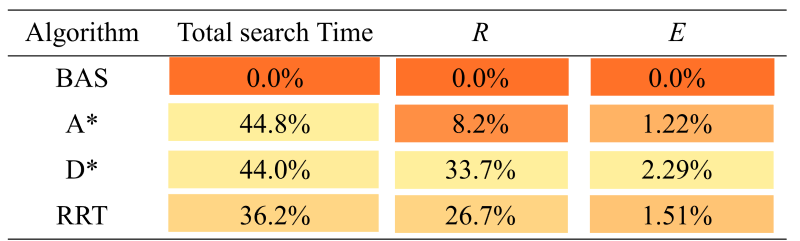}
    \caption{Performance of different methods}
    \vspace{-5mm}
    \label{Performance}
\end{figure}


\section*{Acknowledgment}
This work was supported by the National High Technology Research and Development Program of China under Grant No. 2018YFE0204300, the National Natural Science Foundation of China under Grant No. 62273198, U1964203.

\bibliography{Bibliography}
\bibliographystyle{IEEEtran}
\end{document}